\documentclass[sigconf, screen, nonacm]{acmart}

\AtBeginDocument{%
  }

\setcopyright{none}
\renewcommand\footnotetextcopyrightpermission[1]{}

\usepackage{amsmath}
\usepackage{amsfonts}
\usepackage{booktabs}
\usepackage{graphicx}
\usepackage{pifont}

\newcommand{\cmark}{\textcolor{green!60!black}{\ding{51}}}
\newcommand{\xmark}{\textcolor{red}{\ding{55}}}
\newcommand{\dE}{$\Delta$E}
\newcommand{\dEbf}{\textbf{$\Delta$E}}

\begin{document}
\raggedbottom

\title{Structured-Prior-Guided Diffusion Inpainting with Physical Consistency
for Traffic Sign Augmentation}

\author{Luo Li}
\email{liluo.ll@alibaba-inc.com}
\affiliation{%
  \institution{AMAP, Alibaba Group}
  \city{Beijing}
  \country{China}
}

\author{Chongchong Huang}
\email{huangchongchong.hc@alibaba-inc.com}
\affiliation{%
  \institution{AMAP, Alibaba Group}
  \city{Beijing}
  \country{China}
}

\author{Jun Jia}
\email{tianchen.jj@alibaba-inc.com}
\affiliation{%
  \institution{AMAP, Alibaba Group}
  \city{Beijing}
  \country{China}
}

\author{Qiang Gao}
\email{longmo.gq@autonavi.com}
\affiliation{%
  \institution{AMAP, Alibaba Group}
  \city{Beijing}
  \country{China}
}

\author{Xinlong Liu}
\email{wendiao.lxl@alibaba-inc.com}
\affiliation{%
  \institution{AMAP, Alibaba Group}
  \city{Beijing}
  \country{China}
}

\author{Gui Yang}
\email{yanggui.yg@alibaba-inc.com}
\affiliation{%
  \institution{AMAP, Alibaba Group}
  \city{Beijing}
  \country{China}
}

\author{Liang Cao}
\authornote{Corresponding author.}
\email{xuanhe.xh@autonavi.com}
\affiliation{%
  \institution{AMAP, Alibaba Group}
  \city{Beijing}
  \country{China}
}

\renewcommand{\shortauthors}{Luo Li et al.}

\begin{abstract}
Traffic sign detection faces a long-tailed data distribution. Many rare signs
matter as much as common ones from a regulatory standpoint, yet they have very
few samples. Generative data augmentation is one way out. General-purpose
inpainting models, however, distort digits, deform geometry and perspective,
and shift colours when applied directly to sign regions. We trace this to a
single gap: the conditioning signal is too abstract for the physical
composition of a sign. We propose a structured-prior-guided diffusion
inpainting framework with physical consistency. It injects the semantic,
appearance and geometric priors of a sign through three orthogonal pathways: a
JSON-formatted text prompt, a front-view vector template rendered with measured
dominant colours (via IP-Adapter), and an affine-aligned vector template (via
ControlNet). Two physical consistency losses constrain colour with a CIELAB
chromaticity $L_1$ term and edge structure with a Sobel gradient term. We train
by self-supervised reconstruction on a large set of images collected in-house at
AMAP, then evaluate zero-shot on the public TT100K-2021 dataset, a different
source. Our method uses a Stable Diffusion 1.5 backbone of about 1.4B
parameters. It beats seven representative competitors on every metric of
reconstruction fidelity, physical consistency and semantic controllability. Its
OCR exact-match rate reaches 91.1\%, against 44.2\% for the 12B industrial
model FLUX.1 Fill [dev], and it needs only $1/14$ of that model's inference
time. Leave-one-out ablations confirm that each of the three prior pathways and
both loss terms contribute on their own. In
downstream detection, the synthetic data raises the group-pooled AP50 of rare
classes by $1.23\times$ to $7.40\times$ over a real-data-only baseline. Code and
pre-trained models are available at
\url{https://github.com/52hz-whale/TrafficSignInpaint}.
\end{abstract}

\maketitle

\section{Introduction}

Traffic sign detection and recognition is a core part of autonomous driving
perception and of automated high-definition map production. Large datasets such
as TT100K~\cite{tt100k} and GTSRB~\cite{gtsrb} have appeared, and the
YOLO~\cite{yolo,yolov8,yolo26,yolov10} and DETR~\cite{rtdetr,detr} detector
families have matured. Accuracy on common signs is close to saturated. What
still limits usability is the small set of categories in the tail. This tail
reflects real-world frequency, not sampling bias. More collection mileage buys
very little, and tail categories are usually small objects as well.

Conventional data augmentation has a clear ceiling. Geometric and photometric
transforms and Mosaic~\cite{yolov4} leave semantic content untouched, so they
cannot create new categories. Copy-Paste~\cite{copypaste,cutpaste} and its
Poisson-blending~\cite{poisson} variants move objects around, but the pasted
content still comes from existing instances. Resampling and class-balanced
losses~\cite{cbloss} only reweight the optimisation. They add no information.
Diffusion models~\cite{ddpm,ddim,guided,ldm} and the controllable-generation
ecosystem around them~\cite{controlnet,t2iadapter,ipadapter} open another
route. Suppose we can replace a sign in a real scene with a sign of any chosen
category and content, and the new sign keeps the illumination, occlusion,
background and viewpoint of the original location. Real scenes then become
carriers for long-tailed samples, and the synthetic data inherits the scene
diversity of real data. Recent work synthesises traffic visual elements with
generative models~\cite{raresign} and detection training data with diffusion
models~\cite{xpaste,gen2det,syntheticready,dafusion,datasetdm}. A
general-purpose inpainting model applied directly to signs, however, fails in
predictable ways.

Three failure modes appear, and they are entangled. The first is distortion of
the text on the sign face. The semantic core of a sign is often two or three
characters. ``Speed limit 100'' and ``speed limit 10'' are entirely different
regulatory constraints. Diffusion denoising guarantees nothing about
character-level topological
correctness~\cite{textdiffuser,anytext,glyphcontrol}. A wrong digit is label
noise. It does more harm than generating nothing at all. The second is
destruction of geometry and perspective. The shape of a sign carries semantics
and obeys perspective projection. A circular sign appears as an ellipse whose
axis ratio and tilt follow the viewing angle. The three vertices of a
triangular sign satisfy a specific perspective relation. Natural language
cannot convey these continuous quantities, and generated results come out with
incomplete edges and skewed shapes. The third is colour inconsistency with the
background. The measured colour of one sign differs enormously under harsh noon
sunlight, low-angle dusk light and rainy diffuse light. A model that paints the
standard colours produces a result detached from the surrounding illumination.
For the downstream detector that is a domain shift.

All three have one cause. The conditioning information sits at the wrong level
of abstraction for the target. A traffic sign is a standardised artefact.
National standards prescribe its semantics, colour scheme and geometry in
detail. It carries little information but is strongly structured. Natural
language is a low-precision, ambiguous condition. Driving generation with it
asks the model to guess quantities that are already known.

We define the task as follows. A real road image comes with a mask of the sign
region and a target category label. Structured priors guide a diffusion model
to repaint that region, and the new sign must follow the standard in semantics,
colour and geometry. Our idea is to split the known priors of a sign along
physical dimensions and give each one the conditioning form it fits. Semantics
becomes a structured JSON text prompt. Appearance colour becomes a front-view
vector template infused with the measured dominant colours, injected through
IP-Adapter~\cite{ipadapter}. Geometry becomes an affine-aligned vector
template, injected through ControlNet~\cite{controlnet}. The three pathways are
orthogonal, and none substitutes for another. Together they let the model read
the priors instead of guessing them.

Our main contributions are as follows.

\begin{itemize}
\item \textbf{A structured-prior-guided diffusion inpainting framework.} We
split sign priors into semantics, appearance and geometry, and inject them
through separate pathways. An automated pipeline builds all three conditions
from raw annotations, using SAM~2~\cite{sam2} segmentation, analytic geometric
fitting and colour clustering. Training is self-supervised reconstruction. On
top of the diffusion objective we add a CIELAB chromaticity $L_1$ loss for
colour and a Sobel gradient $L_1$ loss for edge structure. At inference,
editing the conditions edits category, digits, colour and geometry.
\item \textbf{Comprehensive leadership in generation quality and
controllability.} Under a unified TT100K evaluation protocol, our method ranks
first on all nine metrics. Structured prior modelling on a vertical task beats
merely scaling up a general-purpose model.
\item \textbf{Substantial gains from synthetic data on rare-class detection.}
We verify this with YOLO26 on TT100K rare classes. Our synthetic data raises
pooled AP50 by $1.23\times$ to $7.40\times$, and the gain reaches $7.40\times$
when only 10\% of the real data is available.
\end{itemize}

\section{Related Work}

\textbf{Image inpainting.} Early inpainting used GAN and Transformer backbones.
LaMa~\cite{lama} enlarges the receptive field with fast Fourier convolutions.
MAT~\cite{mat} improves texture realism with long-range modelling. This
generation of methods is text-free. It can only erase a region and continue the
surrounding texture plausibly. Its edge scores come from that texture
continuation, not from correct reconstruction of the target structure. On the
diffusion side, DDPM~\cite{ddpm}, DDIM~\cite{ddim} and guidance
mechanisms~\cite{guided,cfg} established the conditional generation framework.
Latent diffusion~\cite{ldm} moved denoising into the VAE latent space. Several
text-conditioned inpainting methods followed. Imagen
Editor~\cite{imageneditor} conditions jointly on text and mask shape and
proposes the EditBench benchmark. PowerPaint~\cite{powerpaint} unifies multiple
inpainting modes with learnable task prompts. BrushNet~\cite{brushnet} injects
masked-image features through a parallel UNet branch. ASUKA~\cite{asuka}
replaces text conditioning with a visual prior to suppress hallucinated
insertion and colour inconsistency. FLUX.1 Fill [dev]~\cite{fluxfill} reaches a
very strong level on general inpainting at 12B scale. All of them carry the
conditioning in natural language and reference photographs. None offers a
precise geometric and chromatic constraint channel for standardised artefacts.
Their evaluation targets general perceptual quality and skips task-critical
questions such as whether the digit on the sign face is correct.

\textbf{Conditional injection and reference-driven generation.}
ControlNet~\cite{controlnet} injects spatially aligned conditions such as edge
and depth maps. It duplicates the encoder of the diffusion backbone and
connects it with zero-initialised convolutions, and it is the de facto standard
for spatially controllable generation. T2I-Adapter~\cite{t2iadapter} reaches a
similar goal with a lighter adapter. IP-Adapter~\cite{ipadapter} targets
appearance conditions that are not spatially aligned. It injects
CLIP~\cite{clip} image embeddings into the UNet through decoupled
cross-attention. This split is the design rationale of our framework.
Geometric shape needs a strong spatially aligned constraint. Colour scheme and
style need a soft non-spatial one. Reference-driven object insertion takes
another line. Paint-by-Example~\cite{paintbyexample} replaces masked content
with a reference image. AnyDoor~\cite{anydoor} performs zero-shot object
teleportation with dual identity and detail pathways. These methods use a real
photograph as the reference. We use a standard vector template instead, and two
differences follow. The template is a noise-free, illumination-free canonical
representation, so it imports no illumination or background from a reference
photograph. The template is also generated programmatically from a category
label, so it covers zero-shot categories just as well. FLUX.1
Kontext~\cite{kontext} and Qwen-Image-Edit~\cite{qwenimage} represent the
instruction-editing paradigm. They edit with natural language and need no mask.
The model itself decides the editing extent, so confining the edit strictly to
the sign region is hard to guarantee.

\textbf{Visual text generation.} The digits on the sign face must be correct,
which ties this task to visual text generation.
TextDiffuser~\cite{textdiffuser} guides text rendering with character-level
segmentation masks. It also established the convention of scoring generated
text with OCR metrics. GlyphControl~\cite{glyphcontrol} injects glyph images as
a spatial condition. AnyText~\cite{anytext} combines glyph, position and mask
for multilingual text generation and editing. We share the idea of using glyphs
or templates as a spatial condition. Our conditioning form differs in two
respects. First, regulation enumerates a tiny character set for traffic signs,
so general glyph-rendering capacity is not what we need. Second, the text on a
sign and its carrier are inseparable. Text, colour and geometry therefore live
in one and the same vector template, not in a separate text layer. We follow
this field in adopting character accuracy and exact-match rate, with
PaddleOCR~\cite{ppocr} as the evaluator.

To the best of our knowledge, no prior work injects semantic, appearance and
geometric priors explicitly and adds loss-level physical consistency
constraints on traffic signs. Nor does any prior work trace a complete evidence
chain from generation quality to downstream detection.

\begin{figure*}[t]
\centering
\includegraphics[width=\textwidth]{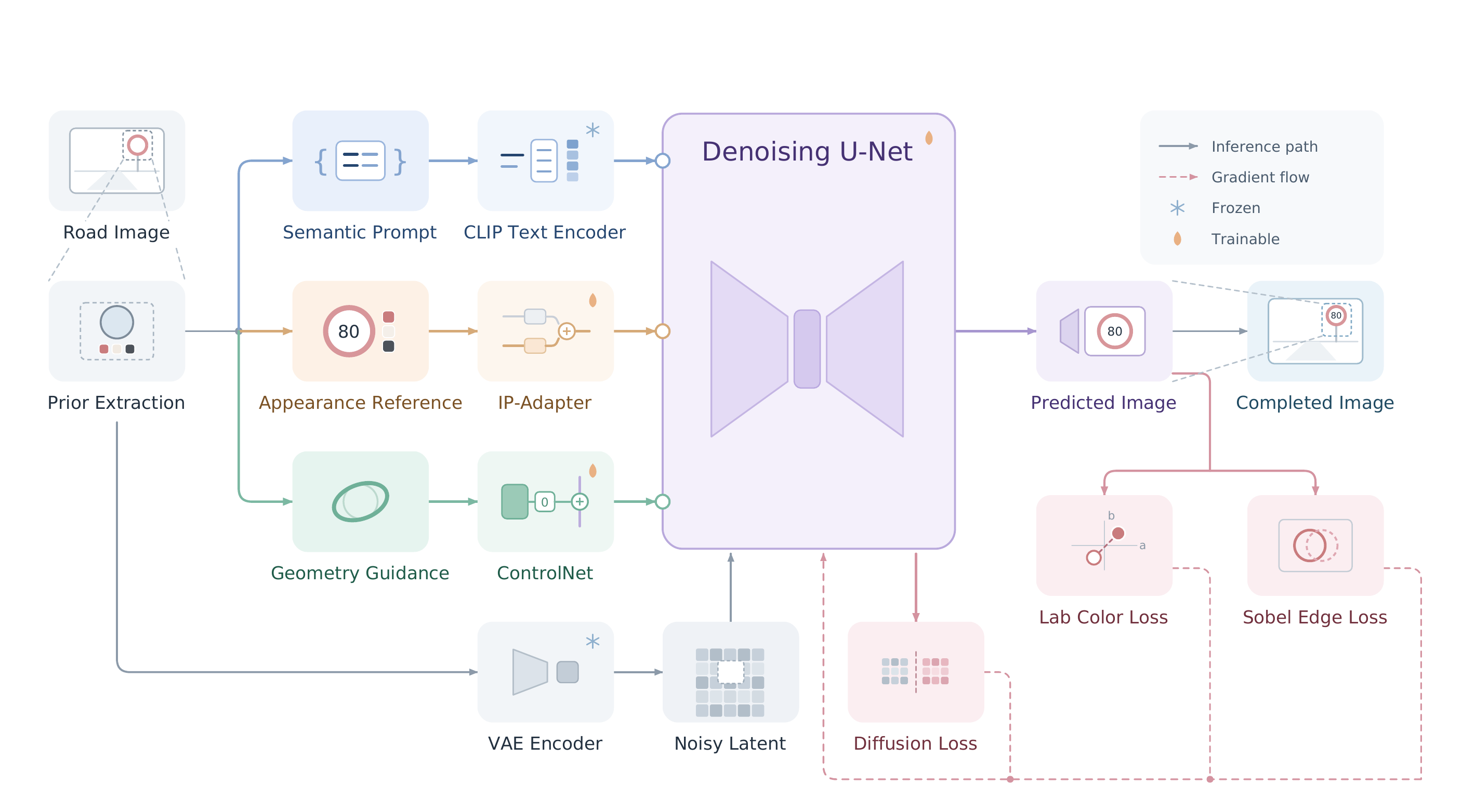}
\caption{Overview of the framework. The semantic (JSON prompt), appearance
(IP-Adapter) and geometric (ControlNet) conditions are injected into the SD1.5
inpainting UNet.}
\Description{Overall framework with three structured conditioning pathways}
\label{fig:framework}
\end{figure*}

\section{Method}

\subsection{Task Definition and Overall Framework}

\textbf{Formalisation.} Given a road image $I \in \mathbb{R}^{H \times W \times 3}$,
a binary mask $M \in \{0,1\}^{H \times W}$ of the sign region, and a structured
description $y = (\text{type}, \text{info})$ of the target sign (for example
type = maximum speed limit, info = 100), the task is to generate an image
$\hat{I}$ satisfying
\begin{equation}
\hat{I} \odot (1-M) = I \odot (1-M), \qquad
\hat{I} \odot M \sim p(\text{sign} \mid y, \mathcal{C}),
\label{eq:task}
\end{equation}
Pixels outside the mask stay unchanged. Pixels inside the mask follow the sign
distribution, conditioned on the structured description $y$ and a derived
condition set $\mathcal{C}$. We split the condition set into three orthogonal
components,
\begin{equation}
\mathcal{C} = \{ c_{\text{sem}},\; c_{\text{app}},\; c_{\text{geo}} \},
\label{eq:cond}
\end{equation}
where $c_{\text{sem}}$ is the JSON-formatted semantic prompt text,
$c_{\text{app}}$ is the front-view vector template infused with the measured
dominant colours, and $c_{\text{geo}}$ is the vector template affine-aligned to
the image pose. Each enters the denoising network by its own route.
$c_{\text{sem}}$ goes through cross-attention after the text encoder.
$c_{\text{app}}$ goes through IP-Adapter decoupled cross-attention after the
CLIP image encoder. $c_{\text{geo}}$ goes through residual injection by the
ControlNet branch.

\textbf{Training paradigm.} During training we extract all
conditions automatically from the original sign region, and the
reconstruction target is the original sign itself. No manually paired data is
needed. Any road image with a sign annotation becomes a training sample.

\textbf{Inference paradigm.} At inference, changing the
conditions changes the generated content. Replacing type/info in $y$ edits
category and digits, which is how we supplement long-tail categories and
generate rare classes. Replacing the clustered colours edits hue and lightness.
Replacing the geometric parameters edits rotation and scale. The three pathways
are decoupled, so these edits apply on their own or together.

\textbf{Overall framework.} Figure~\ref{fig:framework} shows the three steps.
The mask and condition construction module takes the original image and
produces the repaint mask, the shape mask and the three conditions. The three
conditions then drive SD1.5 Inpainting together with ControlNet to repaint the
region. The repainted result is pasted back into the original-resolution image
according to the crop box, ready for downstream detection training. Training
and inference share the same pathway. They differ in two places only. Training
draws the conditions from the original sign region and applies the two physical
consistency losses to $\hat{x}_0$ at low-noise steps. Inference rebuilds the
conditions from the target category.

\begin{figure*}[t]
\centering
\includegraphics[width=\textwidth]{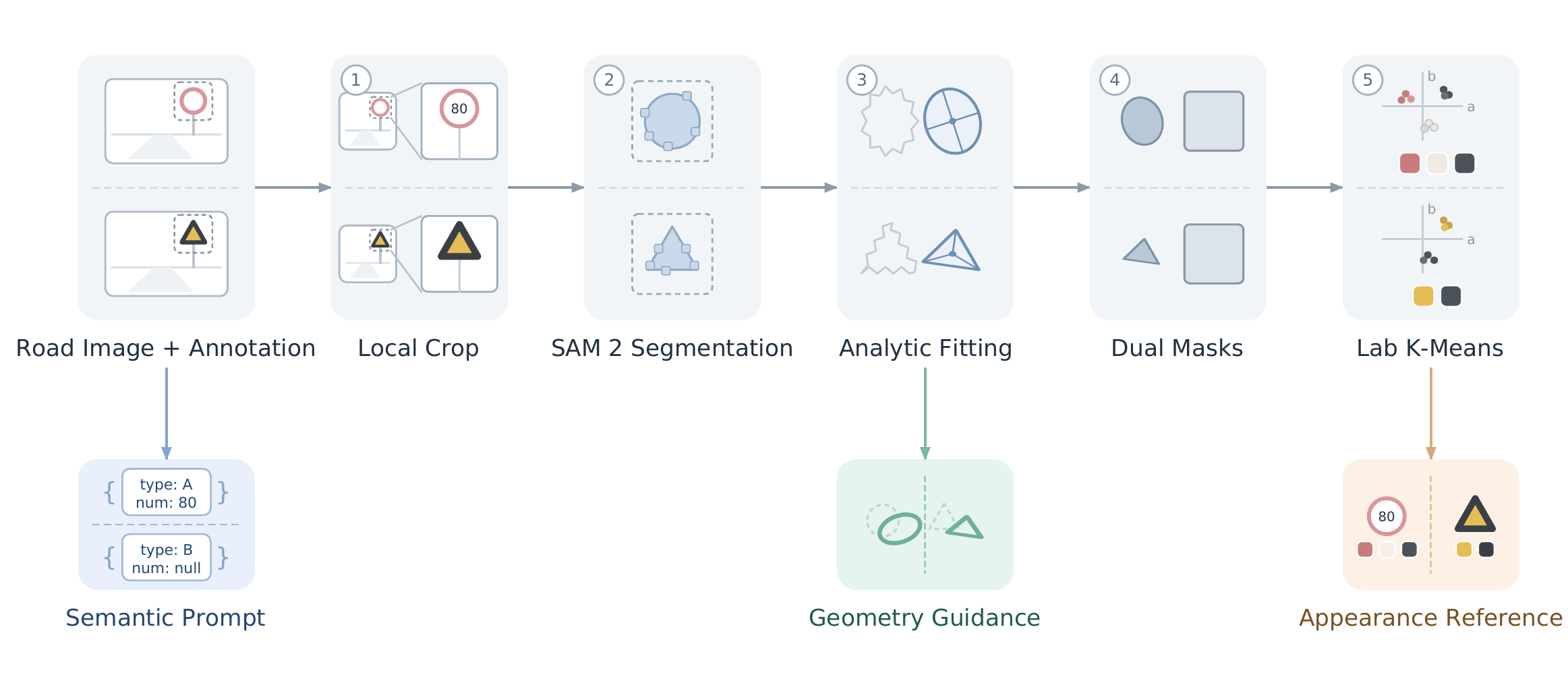}
\caption{Data construction pipeline: local cropping, SAM~2 segmentation,
analytic geometric fitting, dual mask construction and Lab dominant colour
clustering, all fully automatic.}
\Description{Five-step automated data construction pipeline}
\label{fig:pipeline}
\end{figure*}

\subsection{Data Construction Pipeline and Mask Strategy}

Figure~\ref{fig:pipeline} shows the data construction pipeline. It has five
sub-steps, all automated, with no manual annotation involved.

\textbf{(1) Local cropping.} We crop a square region from the original image,
centred near the sign annotation box. Its side length is 3$\times$ the long side
of the annotation box. The crop position is random, subject to the annotation
box keeping a margin from the crop boundary. All crops are resampled to
512$\times$512.

\textbf{(2) SAM~2 segmentation.} The segmentation model is SAM~2~\cite{sam2},
the second generation of SAM~\cite{sam}, prompted with the annotation box to
obtain a pixel-level mask of the sign.

\textbf{(3) Analytic geometric fitting.} We fit per geometric family. For
circular signs we fit an ellipse on the largest contour, which gives the centre,
the two semi-axes and the tilt angle. For triangular signs we run probabilistic
Hough line detection on the mask edge, then enumerate triples of lines and
intersect them to obtain candidate triangles. We score each fit by its IoU with
the SAM mask and discard the samples that fail.

\textbf{(4) Dual mask construction.} The repaint mask $M_{\text{inpaint}}$ is a
rectangle, obtained by dilating each of the four sides of the annotation box by
5 pixels. It is the inpainting input of the diffusion model. The shape mask
$M_{\text{shape}}$ fills the analytic fitting result. It bounds the colour
clustering statistics and the domain of the physical consistency losses.

\textbf{(5) Lab dominant colour clustering.} We run $K$-means in Lab colour
space over the pixels inside the shape mask. $K$ is the number of colours in the
national-standard colour scheme of that category. The standard colour names must
match the cluster centres one to one, otherwise we discard the sample. For
speed-limit signs we keep the pixel coordinates of the cluster holding the
digits. The tilt angle of their minimum bounding rectangle is the sign-face
rotation angle $\theta$.

\subsection{Structured Prior Injection Modules}

\textbf{JSON-formatted semantic prompt.} Instead of free-form natural language
descriptions, we express sign semantics with a fixed JSON structure:
\begin{verbatim}
{"type": "maximum speed limit", "info": "100"}
{"type": "end of speed limit",  "info": "70"}
{"type": "merge from right (triangle)", "info": ""}
\end{verbatim}
There are three reasons behind structured text. First,
it removes ambiguity and redundancy. In natural language ``limit
one hundred'', ``100 km/h speed limit sign'' and ``maximum speed 100'' carry
the same meaning, yet the text encoder maps them to different embeddings.
Second, it is enumerable. The Cartesian product of the category table and the
value table lists every legal condition, so we pre-compute and cache all text
embeddings offline and read them at zero cost during training. Third, it is easy
to edit. Changing one field at inference completes a category or digit edit,
with no sentence to rewrite.

\begin{figure*}[t]
\centering
\includegraphics[width=\textwidth]{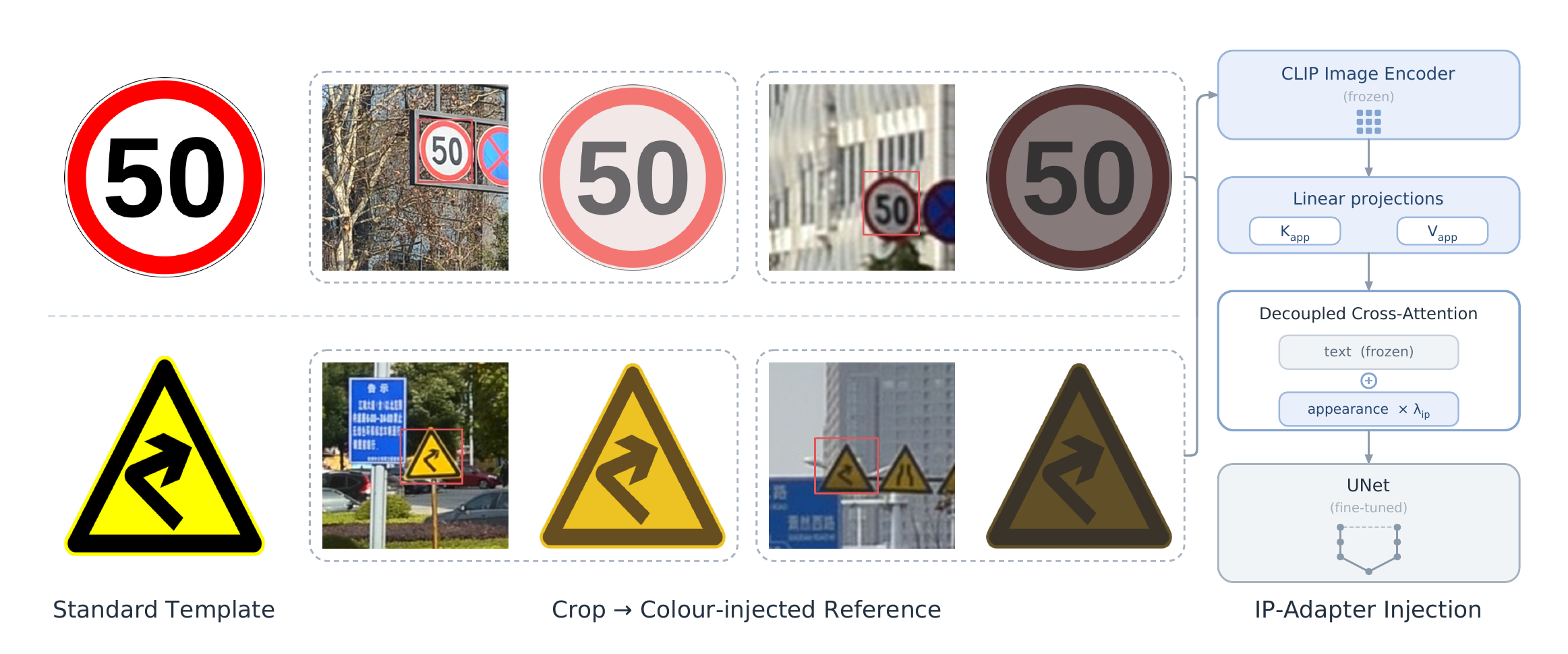}
\caption{Construction of the IP-Adapter appearance prior. Left: the
national-standard vector templates. Middle: the local crops and the reference
images after infusing the measured dominant colours. Right: the path by which
the reference image is injected into the UNet.}
\Description{IP-Adapter appearance prior construction}
\label{fig:appearance}
\end{figure*}

\textbf{IP-Adapter appearance guidance.} The appearance prior rides on a
front-view vector template infused with the measured dominant colours.
Figure~\ref{fig:appearance} shows the construction. The category label indexes
the national-standard vector graphic (SVG). The measured dominant colours
replace the standard colour definitions in that SVG. The result is rasterised
into a 512$\times$512 reference image. Measured colour schemes for one category
differ markedly across illumination conditions, so the reference image follows
the scene instead of being a fixed per-category template.

The CLIP image encoder~\cite{clip} turns this reference image into an image
embedding, and the decoupled cross-attention of IP-Adapter~\cite{ipadapter}
injects it into the UNet. Each UNet layer gains an image pathway alongside the
text pathway. The two share the query and keep independent keys and values,
\begin{equation}
\mathrm{Attn}(Q,K,V) + \lambda_{\text{ip}}\,
\mathrm{Attn}(Q,K_{\text{app}},V_{\text{app}}),
\label{eq:ipa}
\end{equation}
where $K_{\text{app}}, V_{\text{app}}$ are obtained from the CLIP image
embedding of the reference image $c_{\text{app}}$ through independent projection
matrices, and $\lambda_{\text{ip}}$ is the IP-Adapter scale in the inference
configuration. We choose IP-Adapter over a spatial conditioning pathway because
appearance is a global attribute. Colour scheme, style and material feel are not
aligned to pixel positions and should not be bound to them.

\begin{figure*}[t]
\centering
\includegraphics[width=\textwidth]{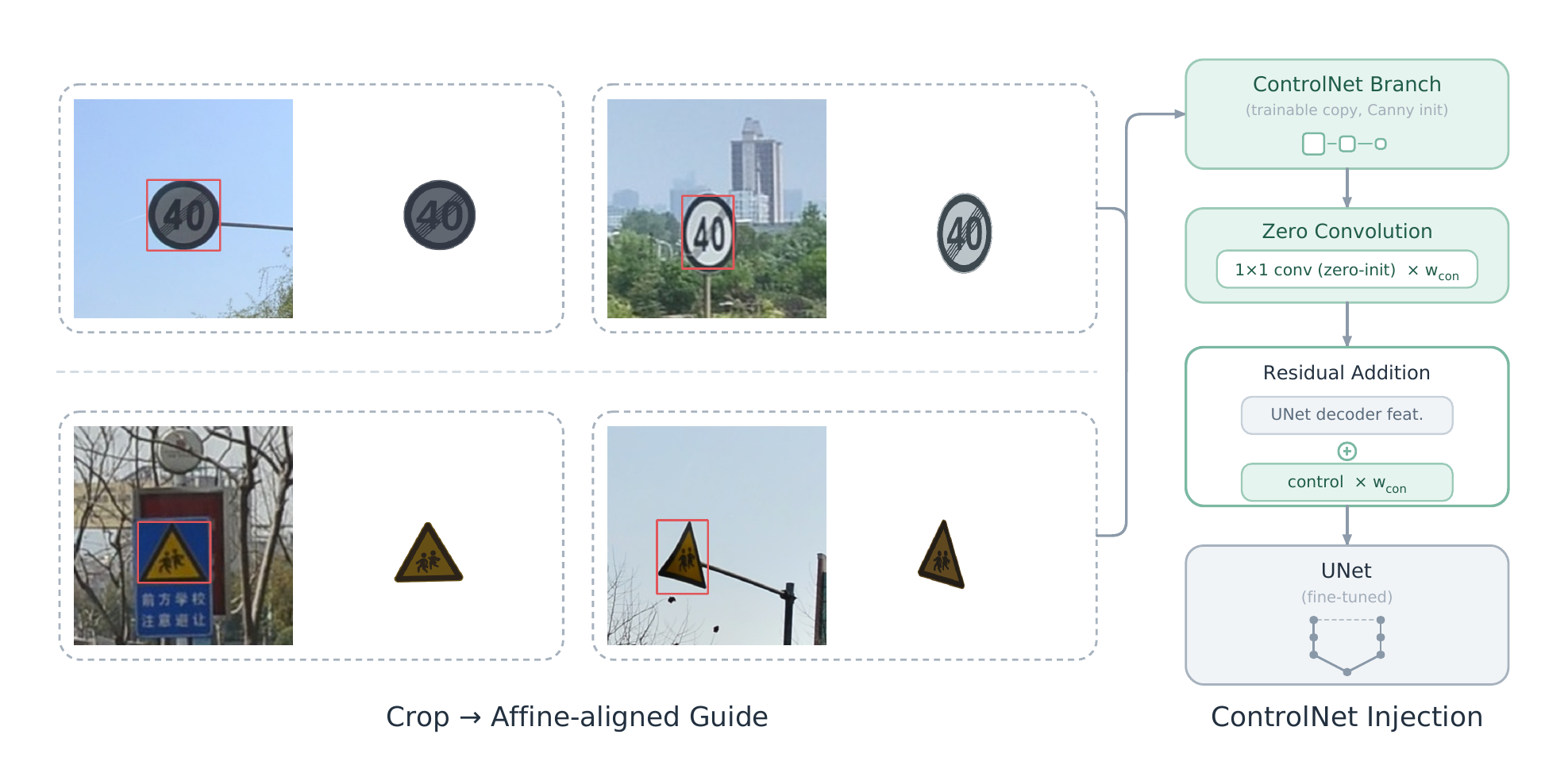}
\caption{Construction of the ControlNet geometric prior. Left: the local crops
and the colour template guidance images affine-aligned according to the analytic
geometric parameters. Right: the path by which the ControlNet branch is added to
the UNet.}
\Description{ControlNet geometric prior construction}
\label{fig:geometry}
\end{figure*}

\textbf{ControlNet geometric control.} The geometric prior rides on the vector
template, affine-transformed into alignment with the image pose.
Figure~\ref{fig:geometry} shows guidance images for one category under a
near-front-view pose and a strongly perspective pose. Both aligned results fit
their pose precisely. The transformed template image is the conditional input of
ControlNet. It enters the features of UNet layer $l$ as
\begin{equation}
h_l' = h_l + Z_l\big(\mathcal{F}_{\text{cn}}(c_{\text{geo}}, z_t, t,
c_{\text{sem}})\big),
\label{eq:cn}
\end{equation}
where $\mathcal{F}_{\text{cn}}$ is the ControlNet branch and $Z_l$ is a
zero-initialised convolution whose scaling factor is the ControlNet
conditioning scale in the inference configuration. One choice here departs from
common usage. We feed the colour vector template image directly as the
condition, not its edge map. The ControlNet branch starts from a Canny-edge
ControlNet and is then fine-tuned end-to-end with the UNet to adapt to this new
conditioning modality. In practice this works better than feeding edge maps. The
template image carries edge positions, region filling relations and the colour
scheme at once, so it holds strictly more information than an edge map.
End-to-end fine-tuning is enough for the branch to exploit that extra
information.

\subsection{Physical Consistency Loss Design}

\textbf{Base diffusion loss.} We adopt the $\epsilon$-prediction MSE loss with
min-SNR-$\gamma$ weighting~\cite{minsnr}:
\begin{equation}
\mathcal{L}_{\text{base}} = \mathbb{E}_{t,\epsilon}\left[
\frac{\min(\mathrm{SNR}(t), \gamma)}{\mathrm{SNR}(t)}
\left\| \epsilon_\theta(\tilde{z}_t, t, \mathcal{C}) - \epsilon \right\|^2
\right], \quad \gamma = 5,
\label{eq:base}
\end{equation}
where $z_t$ is the 4-channel noisy latent at timestep $t$. $\tilde{z}_t$ is the
9-channel UNet input, formed by concatenating $z_t$ with the downsampled mask
and the masked-image latent. The ControlNet branch and the inversion to
$\hat{x}_0$ below both operate on $z_t$. Min-SNR weighting caps the oversized
gradient weights at low-noise steps and balances the effective learning rate
across timesteps. It is known to speed up convergence and improve detail quality
in inpainting fine-tuning.

\textbf{Lab chromaticity consistency loss.} The colour constraint is imposed on
the chromaticity channels of CIELAB space. Let $\hat{x}_0$ be the predicted
image obtained by inverting the model-predicted noise into a clean latent and
decoding it with the VAE, $x$ the ground-truth image, $M_{\text{shape}}$ the
shape mask, and $\mathrm{Lab}(\cdot)_{ab}$ the operation taking the $a,b$
chromaticity channels of Lab. Then
\begin{equation}
\mathcal{L}_{\text{lab}} = \frac{\sum \left| \mathrm{Lab}(\hat{x}_0)_{ab} -
\mathrm{Lab}(x)_{ab} \right| \odot M_{\text{shape}}}
{\sum M_{\text{shape}} + \varepsilon}.
\label{eq:lab}
\end{equation}
Dropping the lightness channel $L$ is deliberate.
Lightness carries illumination, shadow and highlight. Much of its per-pixel
difference is a legitimate variation of scene illumination, and constraining it
would push the model to overfit the illumination of the training samples.
Chromaticity $(a,b)$ stays closer to the intrinsic colour of the object itself,
and that is what the sign standard prescribes. The split also matches the
original intent of Lab space, which separates lightness from chromaticity.

\textbf{Sobel structural consistency loss.} The structural constraint is imposed
on the lightness channel of Lab as the $L_1$ distance between Sobel gradient
magnitudes:
\begin{equation}
\mathcal{L}_{\text{sobel}} = \frac{\sum \left| \;
\mathrm{Sobel}\big(\mathrm{Lab}(\hat{x}_0)_{L}\big) -
\mathrm{Sobel}\big(\mathrm{Lab}(x)_{L}\big) \; \right| \odot M_{\text{shape}}}
{\sum M_{\text{shape}} + \varepsilon}.
\label{eq:sobel}
\end{equation}
The two losses are complementary. The colour constraint uses chromaticity, the
structural constraint uses lightness. They occupy different channels and do not
interfere.

The total loss is
\begin{equation}
\mathcal{L} = \mathcal{L}_{\text{base}} + \lambda_{\text{lab}}
\mathcal{L}_{\text{lab}} + \lambda_{\text{sobel}} \mathcal{L}_{\text{sobel}}.
\label{eq:total}
\end{equation}

\subsection{Training Strategy and Implementation Details}

\textbf{Backbone and trainable modules.} The backbone is the SD1.5 inpainting
checkpoint (\texttt{stable-\allowbreak diffusion-\allowbreak v1-5/\allowbreak
stable-\allowbreak diffusion-\allowbreak inpainting}, a UNet with 9-channel
input). The ControlNet branch starts from
\texttt{lllyasviel/\allowbreak sd-\allowbreak controlnet-\allowbreak canny}. The
trainable parameters are
all weights of the UNet and of ControlNet. The VAE, the CLIP text
encoder and the CLIP image encoder stay frozen throughout. IP-Adapter uses the
SD1.5 version of \texttt{h94/IP-Adapter}, and we fine-tune its image projection
layer and decoupled cross-attention weights together with the UNet.

\textbf{Optimisation configuration.} 8-bit AdamW~\cite{bit8opt,adamw}
($\beta=(0.9,0.999)$, weight decay $10^{-2}$), constant learning rate
$1\times10^{-5}$, gradient clipping threshold 1.0, fp16 mixed precision (the
two-loss path is forced to fp32), 4$\times$V100 data parallelism, effective
batch size 16, and 50{,}000 training steps (about 800k samples). All ablation
variants use exactly the same protocol and number of steps, changing only the
module being ablated.

\textbf{Loss weights.} The weights are calibrated by magnitude
matching. At low-noise steps the base term is about 0.01--0.05 while the two
auxiliary losses are about 2.7 and 1.1. A weight of $3\times 10^{-3}$ therefore
puts their contribution on the same order of magnitude as the base term without
overwhelming it. The two physical losses are evaluated only at low-noise
timesteps ($t < 200$ out of 1000 total steps). There the $\hat{x}_0$ inverted
from $\epsilon_\theta$ has low variance, so the pixel-level constraints act on a
meaningful signal. This also caps the extra VAE decoding cost at roughly 20\% of
the samples.

\textbf{Inference configuration.} Sampler UniPC~\cite{unipc} with 20 steps;
classifier-free guidance scale 3.0; IP-Adapter scale 1.0; ControlNet
conditioning scale 0.75; resolution 512$\times$512; random seed fixed to 42.
The hyper-parameter sensitivity analysis in Section~\ref{sec:quality} selected
this configuration on the Pareto frontier. It sits at a balanced position with
no obvious weakness.

\textbf{Convergence.} The training loss enters a plateau after 10{,}000 steps.
The interval mean of the total loss falls from 0.0173 over steps
10{,}000--15{,}000 to 0.0163 over steps 45{,}000--50{,}000, a 5.8\% reduction.
Over the same period the Lab and Sobel terms fall by 3.1\% and 2.9\%.
Training for 50{,}000 steps has thus entered the convergence regime, and further
steps buy little.

\section{Experiments}

\subsection{Datasets, Evaluation Protocol and Implementation}

\textbf{Training set.} The training data is road
images collected in-house, paired with their sign annotations (category +
annotation box + digits). There are 1{,}673{,}112 pairs. They cover 41
templates: 19 triangular warning classes and 22 circular
speed-limit classes with different digits.

\textbf{Evaluation set.} We evaluate on TT100K-2021.
Its collection source differs entirely from that of the training data. We read
all TT100K annotations and keep the targets in the 41 supported categories,
which gives 7{,}898 candidates. These pass three automatic gates: SAM~2
segmentation, analytic geometric fitting and one-to-one colour-scheme matching.
5{,}182 sign regions remain as the evaluation set, and 4{,}601 of them are
speed-limit classes bearing digits. This dataset is also heavily long-tailed and
dense in small objects. The 9{,}050 images left after excluding the
\texttt{other} split hold 25{,}010 target instances across 184 categories. 83 of
those categories have a single-digit number of instances, and 46.3\% of targets
have a short side below 32 pixels.

\textbf{Evaluation metrics.} Metrics are chosen along three dimensions.
Reconstruction fidelity: PSNR/SSIM~\cite{ssim}, LPIPS~\cite{lpips} and
FID~\cite{fid} (Inception-v3~\cite{inception} features). Physical consistency:
CIEDE2000 colour difference~\cite{ciede2000} and in-mask IoU/F1 on
Canny~\cite{canny} edge maps. CIEDE2000 is a perceptually uniform industrial
standard with a discrimination threshold of \dE$\approx$5, and it beats a simple
RGB distance. Semantic controllability: OCR exact-match
rate (EM), character accuracy (CharAcc) and normalised edit distance (NED),
with PaddleOCR~\cite{ppocr} as the evaluator.

\textbf{Competing models.} There are seven, all run zero-shot with official
pretrained weights. Table~\ref{tab:baselines} lists the role each plays in our
argument.

\begin{table}[t]
\centering
\caption{The seven competing models and the role each plays in the comparison.}
\label{tab:baselines}
\small
\begin{tabular}{@{}lp{4.05cm}@{}}
\toprule
\textbf{Model (venue)} & \textbf{Role in our argument} \\
\midrule
LaMa (WACV'22)~\cite{lama} & Non-diffusion (FFC-GAN) generational anchor \\
MAT (CVPR'22)~\cite{mat} & Non-diffusion (Transformer) large-hole anchor \\
SD1.5 Inpainting (2022)~\cite{ldm} & Bare lower bound of our own backbone \\
PowerPaint v2 (ECCV'24)~\cite{powerpaint} & Task-prompt style multi-task inpainting \\
BrushNet (ECCV'24)~\cite{brushnet} & Academic benchmark of the controllable-injection school \\
ASUKA-SD1.5 (CVPR'25)~\cite{asuka} & Fairest rival: same backbone, colour-consistency theme \\
FLUX.1 Fill [dev] (2024)~\cite{fluxfill} & 12B industrial SOTA, parameter-count upper bound \\
\bottomrule
\end{tabular}
\end{table}

Two instruction-editing methods, FLUX.1 Kontext~\cite{kontext} and
Qwen-Image-Edit~\cite{qwenimage}, stay out of the quantitative
comparison. They take no mask input, the model itself decides the editing
extent, and they cannot guarantee that pixels outside the mask remain unchanged,
which data augmentation requires.

\textbf{Fairness protocol.} (i) The same 5{,}182 (image, mask) inputs; (ii) the
same natural-language prompt transcribed from one set of JSON annotations
(except LaMa/MAT, which have no text interface); (iii) fixed random seed 42;
(iv) each method uses its officially recommended inference configuration; (v)
all metrics are computed only inside the mask region, by one and the same
evaluation code.

One difference deserves note. Our method is domain fine-tuned, the competing
models are zero-shot. The same-backbone ablation in Section~\ref{sec:ablation}
covers this difference.

\subsection{Generation Quality Evaluation}
\label{sec:quality}

\begin{table*}[t]
\centering
\caption{Zero-shot comparison against the seven competing models on the
5{,}182-region TT100K evaluation set. Best value per column in bold.}
\label{tab:main}
\small
\begin{tabular}{@{}lccccccccc@{}}
\toprule
\textbf{Method} & \textbf{PSNR}$\uparrow$ & \textbf{SSIM}$\uparrow$ &
\textbf{LPIPS}$\downarrow$ & \textbf{FID}$\downarrow$ & \dEbf$\downarrow$ &
\textbf{EdgeIoU}$\uparrow$ & \textbf{EdgeF1}$\uparrow$ &
\textbf{OCR-EM}$\uparrow$ & \textbf{NED}$\uparrow$ \\
\midrule
LaMa (WACV'22)~\cite{lama}            & 13.59 & 0.508 & 0.112 & 31.70 & 17.40 & 0.086 & 0.127 & 0.2\%  & 0.021 \\
MAT (CVPR'22)~\cite{mat}              & 13.03 & 0.459 & 0.109 & 29.48 & 18.29 & 0.070 & 0.107 & 0.0\%  & 0.019 \\
SD1.5 Inpainting~\cite{ldm}           & 10.51 & 0.346 & 0.123 & 21.25 & 23.36 & 0.051 & 0.095 & 20.7\% & 0.425 \\
PowerPaint v2 (ECCV'24)~\cite{powerpaint} & 9.52 & 0.292 & 0.136 & 24.35 & 26.11 & 0.046 & 0.086 & 23.2\% & 0.455 \\
BrushNet (ECCV'24)~\cite{brushnet}    & 11.16 & 0.361 & 0.125 & 19.06 & 22.05 & 0.048 & 0.090 & 13.9\% & 0.315 \\
ASUKA-SD1.5 (CVPR'25)~\cite{asuka}    & 13.54 & 0.437 & 0.080 & 18.73 & 17.12 & 0.085 & 0.138 & 0.8\%  & 0.074 \\
FLUX.1 Fill [dev] (12B)~\cite{fluxfill} & 13.90 & 0.435 & 0.074 & 9.05 & 16.79 & 0.070 & 0.123 & 44.2\% & 0.650 \\
\midrule
\textbf{Ours (SD1.5, $\sim$1.4B)}     & \textbf{15.53} & \textbf{0.534} & \textbf{0.039} & \textbf{4.22} & \textbf{12.40} & \textbf{0.087} & \textbf{0.144} & \textbf{91.1\%} & \textbf{0.927} \\
\bottomrule
\end{tabular}
\end{table*}

\textbf{(1) First place on all nine metrics.} As Table~\ref{tab:main} shows,
our method leads on reconstruction fidelity, physical consistency and semantic
controllability at once. On the semantic dimension the lead is of a different
order (OCR-EM 91.1\% vs.\ 44.2\%, +46.9\,pt).

\textbf{(2) The three classes of competitors fail in different ways, all
pointing to a deficiency in the conditioning form.} The non-diffusion methods
(LaMa/MAT) do respectably on PSNR/SSIM/EdgeIoU, yet their OCR exact-match rate
is near zero. They erase the sign and continue the background texture
plausibly, and their edge scores come from that continuity rather than from
correct reconstruction of the sign structure. The general text-conditioned
diffusion methods (SD1.5/PowerPaint/BrushNet) do generate sign-like content,
with OCR-EM of 13.9\%--23.2\%. But their colour difference reaches 22--26 and
their PSNR falls below the non-diffusion methods, because the generated result
does not match the illumination and geometry of the scene. ASUKA is the most
instructive case. Its LPIPS (0.080) and \dE\ (17.12) are the best in the
SD1.5 family and second only to the 12B FLUX.1 Fill, so its colour-consistency
design works. Its OCR-EM is nonetheless 0.8\%. The MAE visual prior suppresses
hallucinated object insertion, and it costs the model almost all of its text
generation ability.

\textbf{(3) A 1B-scale task-specialised model surpasses a 12B general-purpose
model.} FLUX.1 Fill [dev] is the strongest competitor (FID 9.05, OCR-EM
44.2\%). Its 12B parameters give it far better general generation ability than
the SD1.5-family baselines. Our method still passes it on all nine metrics with
about $1/9$ of the parameters: +46.9 percentage points on OCR-EM and 53\% lower
FID. Our single-sample inference time on one V100 is 8.81\,s, only $1/14$ of the
123.55\,s of FLUX.1 Fill. The judgement is clear. On vertical tasks with
standardised semantics and geometry, feeding the prior to the model in
structured form beats scaling the model up and letting it guess.

\textbf{(4) OCR confidence separation.} OCR confidence separates sharply.
Correct samples average 0.988, incorrect ones 0.328. This gives a
reference-free signal for screening generation quality.

\begin{figure*}[t]
\centering
\includegraphics[width=0.80\textwidth]{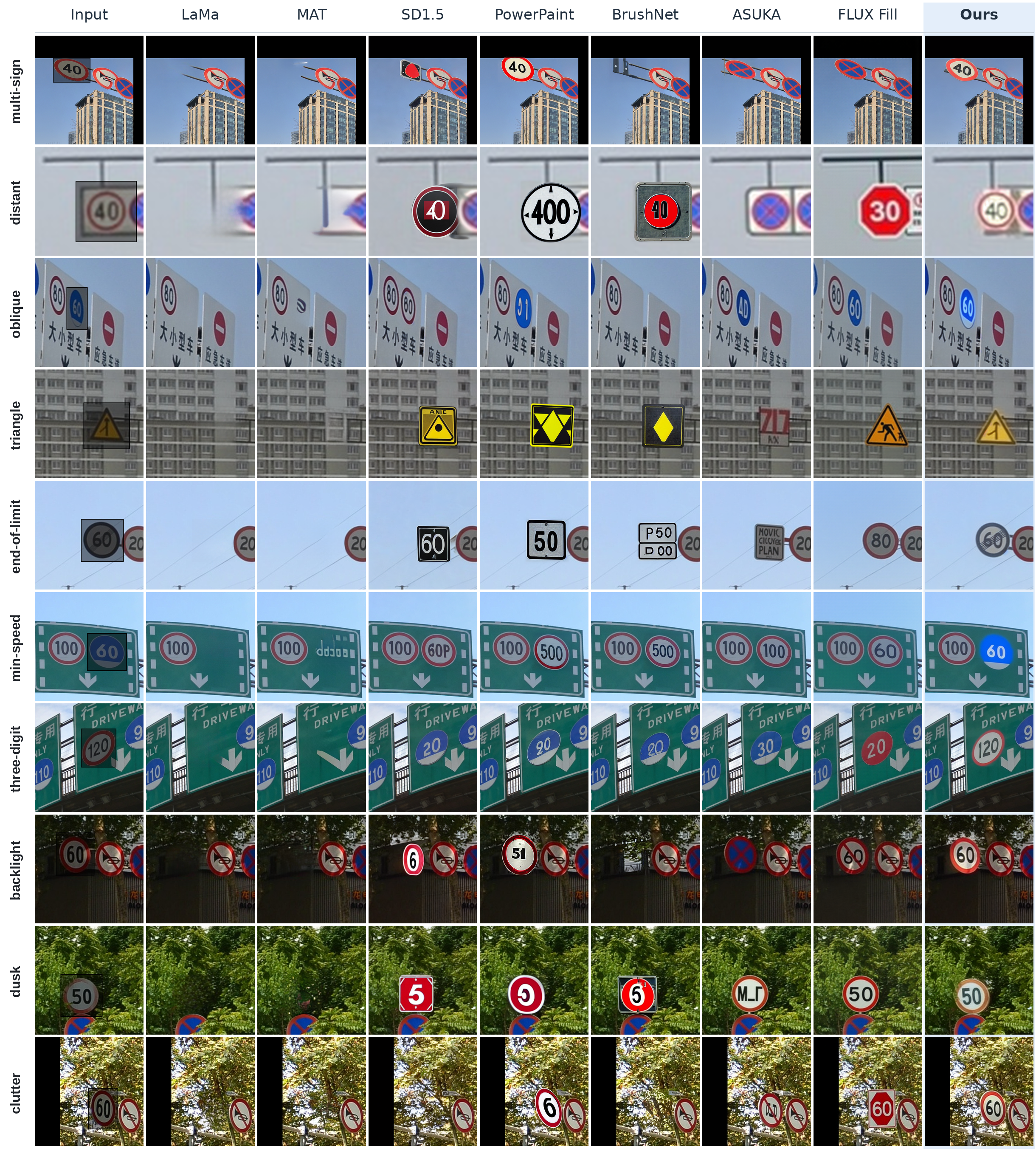}
\caption{Qualitative comparison over ten representative scenes (rows). Each row
shows the input followed by the seven competing models and our result.}
\Description{Qualitative comparison grid over ten scenes, the seven competing
models and our method}
\label{fig:qualitative}
\end{figure*}

\textbf{Cell-by-cell comparison with the baselines.}
Figure~\ref{fig:qualitative} places our method beside the seven competing
models over ten representative scenes, and the trend is clear. The
non-diffusion methods (LaMa/MAT) smooth the sign over with background texture
and lose the sign-face semantics almost entirely. The general text-conditioned
diffusion methods (SD1.5/PowerPaint/BrushNet) produce sign-like content, but
often distort the sign-face geometry and the digit glyphs. FLUX.1 Fill, the
strongest baseline, keeps glyphs far more intact. It still skews the structure
and sticks characters together in strongly perspective and three-digit scenes.
Our method holds the sign outline, the perspective pose and the digit glyphs in
all ten rows.

\begin{figure*}[t]
\centering
\includegraphics[width=\textwidth]{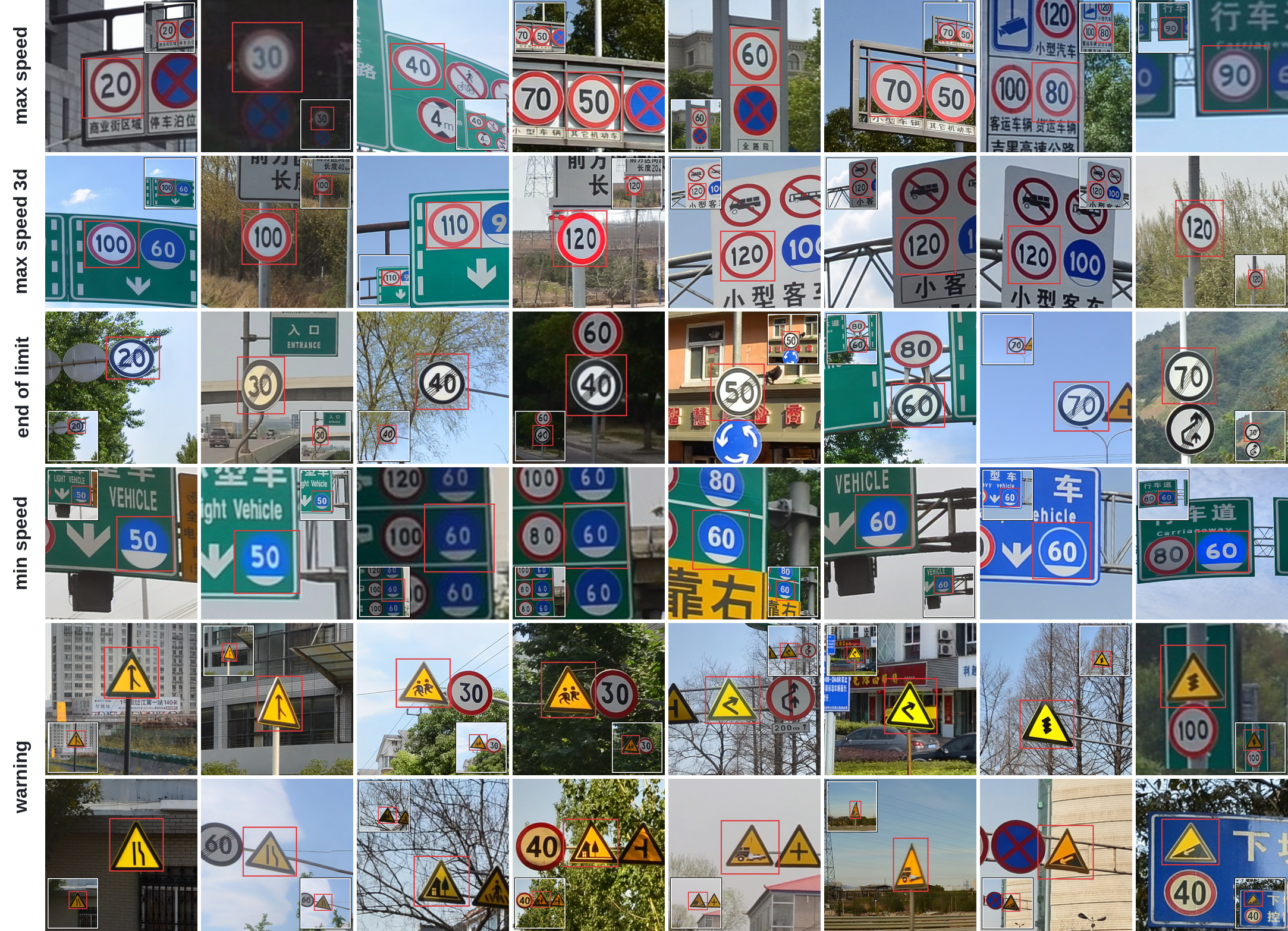}
\caption{Reconstruction quality wall grouped by sign family. Each tile is our
reconstruction with the ground-truth crop inset; the red box marks the repaint
region in both.}
\Description{Reconstruction quality wall with ground-truth thumbnails}
\label{fig:recwall}
\end{figure*}

\textbf{Reconstruction fidelity against the ground truth.}
Figure~\ref{fig:recwall} leaves the baselines aside and examines our
reconstruction fidelity against the ground truth alone. Each tile gives our
reconstruction with the original crop thumbnail overlaid in a free corner as the
reference, and a red box marks the same repaint region on both. Reading the
pairs cell by cell, the sign-face colour scheme, outline and digit glyphs all
agree closely with the ground truth. Night-time low-light and distant
small-scale samples show no obvious degradation either.

\begin{figure*}[t]
\centering
\includegraphics[width=\textwidth]{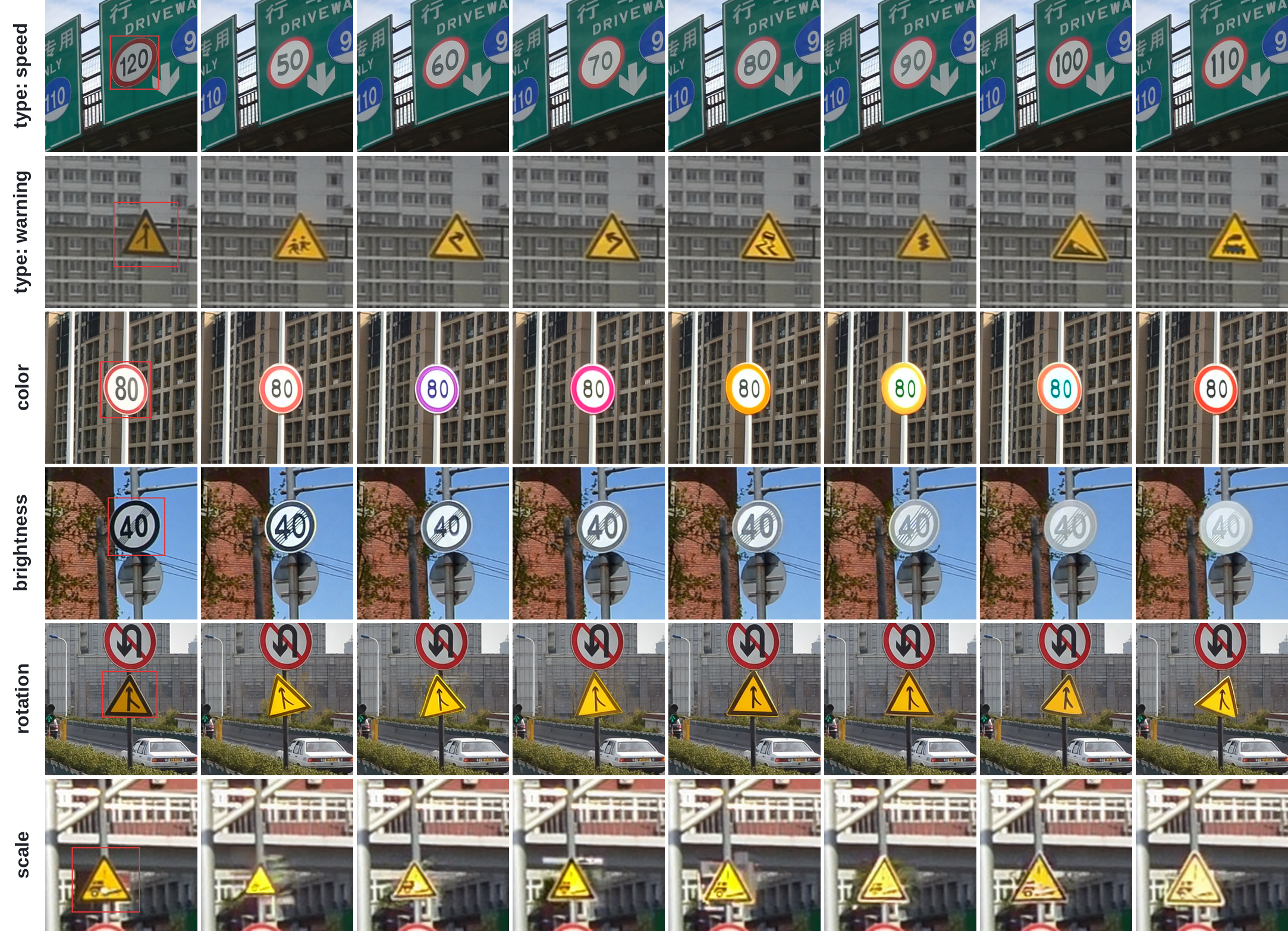}
\caption{Controllability grid. Leftmost column: the input (red box: the repaint
region). Each row edits one condition only: speed-limit value, warning
category, colour scheme, lightness, rotation, scale.}
\Description{Controllability grid over six kinds of single-condition edits}
\label{fig:controllability}
\end{figure*}

\textbf{Independent controllability of the condition pathways.} Here we test the
editability of the condition pathways. We keep the scene and mask fixed, change
one condition component at a time, and check whether the result changes as
intended while the other attributes hold.
Figure~\ref{fig:controllability} shows sequences for six kinds of
edits: changing the speed-limit value, changing the warning sign category,
changing the sign-face colour scheme, and continuous interpolation over
lightness, geometric rotation and scale. In every row the target attribute
varies monotonically and smoothly with the condition. The background pixels and
the unedited attributes stay stable, with no cross-talk between conditions. The
text, appearance and geometry pathways are thus decoupled and individually
effective, which is exactly what a controllable data augmentation engine needs.

\textbf{Sensitivity of inference hyper-parameters.} We selected the sampler and
the three guidance-strength parameters (\texttt{gs}/\texttt{ip}/\texttt{cn}) on a
1000-sample subset with 20 steps. The sampler came first, chosen among UniPC,
DPM++ 2M~\cite{dpmpp} and PNDM. UniPC leads stably by 1.1--2.3\,pt on OCR-EM,
and the LPIPS/FID differences among the three stay below 2\%, so we adopt it
throughout. The guidance parameter grid covers gs $\in \{2.0,\ldots,6.0\}$
$\times$ ip $\in \{0.8, 1.0\}$ $\times$ cn $\in \{0.5, 0.75, 1.0\}$. The Pareto
frontier holds 11 non-dominated solutions, and Table~\ref{tab:pareto} lists the
five core ones.

\begin{table*}[t]
\centering
\caption{The five core groups among the 11 non-dominated configurations on the
Pareto frontier of the inference hyper-parameter grid. The first row is the
adopted configuration; best value per column in bold.}
\label{tab:pareto}
\small
\begin{tabular}{@{}lccccccl@{}}
\toprule
\textbf{Configuration} & \textbf{PSNR}$\uparrow$ & \textbf{SSIM}$\uparrow$ &
\dEbf$\downarrow$ & \textbf{EdgeIoU}$\uparrow$ & \textbf{EdgeF1}$\uparrow$ &
\textbf{OCR-EM}$\uparrow$ & \textbf{Positioning} \\
\midrule
\textbf{gs3.0$\cdot$ip1.0$\cdot$cn0.75} & 15.66 & 0.541 & 12.06 & 0.081 & 0.138 & 90.0\% & Balanced, no weakness \\
gs2.5$\cdot$ip1.0$\cdot$cn0.5 & \textbf{15.74} & 0.525 & \textbf{11.98} & 0.077 & 0.132 & \textbf{90.6\%} & Triple crown: OCR / fidelity / colour \\
gs3.0$\cdot$ip1.0$\cdot$cn0.5 & 15.69 & 0.525 & 12.07 & 0.079 & 0.135 & \textbf{90.6\%} & OCR peak \\
gs2.5$\cdot$ip1.0$\cdot$cn1.0 & 15.53 & \textbf{0.552} & 12.17 & 0.085 & 0.143 & 89.4\% & SSIM champion \\
gs4.0$\cdot$ip1.0$\cdot$cn1.0 & 15.29 & 0.549 & 12.56 & \textbf{0.086} & \textbf{0.145} & 90.2\% & Structure first, OCR retained \\
\bottomrule
\end{tabular}
\end{table*}

Three points follow. The gs knee lies on the 2.5--3.0 plateau. Below it the text
condition is too weak, above it the guidance is too strong.
cn is the core trade-off axis. Moving from 0.5 to 1.0 buys a structural
gain of SSIM +0.027 / EdgeF1 +0.007 at the price of a slight OCR loss.
ip has a mild influence, yet all 11 non-dominated solutions sit at
ip = 1.0. The optimum of the inference hyper-parameters depends on the training
configuration and does not carry over across configurations.
The main table therefore uses the middle of the Pareto frontier:
gs 3.0 / ip 1.0 / cn 0.75 / UniPC / 20 steps.

\subsection{Ablation Study}
\label{sec:ablation}

\textbf{Variant design.} All ablations run on the SD1.5 backbone with
a strict leave-one-out single-variable design. Taking Full as the
reference, we remove exactly one prior module or loss term at a time, which
gives 5 ablation variants plus Full. Table~\ref{tab:ablconfig} lists the
configurations.

\begin{table}[t]
\centering
\caption{Leave-one-out ablation variants. \cmark: module kept; \xmark: module
removed.}
\label{tab:ablconfig}
\small
\setlength{\tabcolsep}{3.2pt}
\begin{tabular}{@{}lcccccc@{}}
\toprule
\textbf{Variant} & \rotatebox{90}{\textbf{JSON Prompt}} &
\rotatebox{90}{\textbf{IP-Adapter}} & \rotatebox{90}{\textbf{Colour inj.}} &
\rotatebox{90}{\textbf{ControlNet}} & \rotatebox{90}{\textbf{Lab Loss}} &
\rotatebox{90}{\textbf{Sobel Loss}} \\
\midrule
V1 Full $-$ IP-Adapter        & \cmark & \xmark & \cmark & \cmark & \cmark & \cmark \\
V2 Full $-$ colour injection  & \cmark & \cmark & \xmark & \cmark & \cmark & \cmark \\
V3 Full $-$ both phys. losses & \cmark & \cmark & \cmark & \cmark & \xmark & \xmark \\
V4 Full $-$ Sobel Loss        & \cmark & \cmark & \cmark & \cmark & \cmark & \xmark \\
V5 Full $-$ ControlNet        & \cmark & \cmark & \cmark & \xmark & \cmark & \cmark \\
V6 Full (Ours)                & \cmark & \cmark & \cmark & \cmark & \cmark & \cmark \\
\bottomrule
\end{tabular}
\end{table}

One clarification is necessary. The measured colours are injected into the vector
template, and both the IP-Adapter reference image and the ControlNet guidance
image derive from that template, so ``colour injection'' is a switch that
spans two condition pathways. V1 removes only the IP-Adapter branch, and the
ControlNet guidance image still carries the measured colours. V2 keeps both
pathways but uses the original template colours throughout. V6 vs.\ V1
therefore isolates the contribution of the appearance/layout condition
pathway, and V6 vs.\ V2 isolates the contribution of measured colour
injection. The two do not overlap.

\begin{table*}[t]
\centering
\caption{Leave-one-out ablation results, V6 being the full model. Best value
per column in bold.}
\label{tab:ablation}
\small
\begin{tabular}{@{}lcccccccccc@{}}
\toprule
\textbf{Variant} & \textbf{PSNR}$\uparrow$ & \textbf{SSIM}$\uparrow$ &
\textbf{LPIPS}$\downarrow$ & \textbf{FID}$\downarrow$ & \dEbf$\downarrow$ &
\textbf{EdgeIoU}$\uparrow$ & \textbf{EdgeF1}$\uparrow$ &
\textbf{OCR-EM}$\uparrow$ & \textbf{CharAcc}$\uparrow$ & \textbf{NED}$\uparrow$ \\
\midrule
\textbf{V6 Full (Ours)}      & 15.53 & 0.534 & 0.039 & 4.22 & 12.40 & 0.087 & 0.144 & 91.1\% & 92.7\% & 0.927 \\
V1 $-$ IP-Adapter            & 16.44 & 0.509 & 0.059 & 16.94 & 11.98 & 0.078 & 0.140 & 66.1\% & 69.6\% & 0.697 \\
V2 $-$ colour injection      & \textbf{16.47} & 0.516 & 0.054 & 9.65 & \textbf{11.65} & 0.079 & 0.142 & 85.1\% & 87.9\% & 0.880 \\
V3 $-$ both phys.\ losses    & 15.45 & \textbf{0.544} & \textbf{0.038} & 3.96 & 12.44 & \textbf{0.090} & \textbf{0.151} & 88.9\% & 90.7\% & 0.907 \\
V4 $-$ Sobel Loss            & 15.01 & 0.517 & 0.042 & 4.27 & 13.13 & 0.081 & 0.137 & 88.9\% & 90.5\% & 0.905 \\
V5 $-$ ControlNet            & 15.54 & 0.513 & 0.040 & \textbf{3.52} & 12.06 & 0.072 & 0.130 & \textbf{92.6\%} & \textbf{93.7\%} & \textbf{0.937} \\
\bottomrule
\end{tabular}
\end{table*}

The following conclusions can be drawn from Table~\ref{tab:ablation}.

\textbf{(1) The IP-Adapter appearance prior is the decisive factor for digit
legibility and distributional realism.} Removing it drops OCR-EM from
91.1\% to 66.1\%, pushes FID from 4.22 to 16.94, and worsens LPIPS by 51\%.
The vector template supplies the colour scheme and the
complete layout structure of the sign face: the glyph shape and
size of the digits, their relative position within the face, and the white
space. Text conditioning cannot convey any of that.

\textbf{(2) Colour injection is the decisive factor for
distributional realism.} V2 keeps IP-Adapter but switches to the original
template colours. FID rises from 4.22 to 9.65, LPIPS worsens by 38\% and
OCR-EM falls by 6.0\,pt. The idealised saturated template colours sit too far
from real imaging colours. The model has to bridge the gap itself, and its
modelling of the other attributes suffers.

\textbf{(3) The ControlNet geometric prior buys structural and fidelity gains at
a slight OCR/FID cost.} Removing ControlNet raises OCR-EM by 1.5\,pt and
lowers FID by 0.70. EdgeIoU drops by 17\%, EdgeF1 by 10\% and SSIM by 0.021.
The trade-off is plain. A strong spatial constraint limits the model's freedom
inside the sign face, and writing the digits correctly needs some freedom. We
keep ControlNet. Geometric correctness matters for downstream detection, for
example the consistency of box shape and pose, and OCR cannot stand in for it.

\textbf{(4) The core benefit of the two physical losses is text accuracy.}
Relative to V3 (no physical losses), OCR-EM is +2.2\,pt while \dE/FID/LPIPS are
essentially unchanged; the cost is SSIM $-$0.010 and EdgeF1 $-$4.6\%.

\textbf{(5) The independent contribution of the Sobel structural loss is a
structural gain at no distributional cost.} Relative to V4 (Lab present, Sobel
absent): EdgeIoU +7.4\%, EdgeF1 +5.1\%, SSIM +0.017, \dE\ $-$0.73 and OCR-EM
+2.2\,pt, while FID is essentially unchanged. The Sobel loss is thus an almost
free gain term.

\textbf{(6) The cost of the two losses on purely structural metrics.} V3 (no
physical losses) keeps a slight advantage over Full on SSIM (+1.9\%), EdgeF1
(+4.9\%), EdgeIoU (+3.4\%) and FID (3.96 vs.\ 4.22). The benefit of the two
losses concentrates in colour fidelity (the \dE\ $-$0.73 isolated by
point (5)) and text accuracy (+2.2\,pt). The cost on purely structural metrics
is real. Point (5) already showed by independent leave-one-out evidence that the
Sobel term brings a structural gain, so the cost here belongs to the indirect
effect of the Lab chromaticity term on lightness structure. Paying 0.010 of
SSIM for 2.2\,pt of OCR is a reasonable deal: for data augmentation, a sample
with a wrong digit is a harmful sample, while a sample with 0.010 lower SSIM is
still a usable one.

\textbf{(7) The failure of PSNR and the ``conservative generation'' trap.} V1 and
V2 reach a PSNR of 16.44 and 16.47, well above the 15.53 of Full. Their FID is
4.0$\times$ and 2.3$\times$ that of Full, and their OCR-EM is 25.0\,pt and
6.0\,pt lower. Without the appearance/colour prior the model generates
conservative content with low contrast, close to the mean of the mask
surroundings. Such content wins on per-pixel squared error and fails completely
in perceptual and semantic terms. The methodological conclusion is clear. On
semantically critical tasks such as sign repainting, PSNR alone cannot evaluate
generation quality. It must be read together with FID/LPIPS/OCR and the
physical consistency metrics.

Full wins no single column in Table~\ref{tab:ablation}. It stays balanced and
near the front on the key dimensions: third on FID, second on OCR behind V5,
second on the Edge metrics behind V3. That is the reasonable landing point of
a multi-objective trade-off.

\subsection{Downstream Detection Verification}

\textbf{Detector and data strategies.} The detector is
YOLO26s-p2~\cite{yolo26} with a P2 detection head (stride 4) for the
small objects of this task. It is only a fixed measuring instrument here. All 16
runs share an identical training protocol, item by item. Horizontal flipping is
disabled: the category set contains three pairs of left--right mirrored classes,
and flipping would manufacture exact mislabels. The
four data strategies are: (i) real data only (baseline); (ii) real +
Copy-Paste hard paste; (iii) real + Copy-Paste Poisson blending~\cite{poisson};
(iv) real + our diffusion synthesis.

\textbf{Category protocol and main metric.} The 34 classes are 14 Tier-1
classes, 19 Tier-2 classes and an \texttt{other\_sign} catch-all. The
capabilities the diffusion model already has determined this set in reverse.
Data come from the official TT100K-2021 train split, divided into train/val, and
we evaluate on test exactly once. The \texttt{other} split is excluded entirely
because its annotation is incomplete: missing labels would count as false
positives and systematically depress AP. The main metric is
Tier-2 in-group pooled AP50. We merge the 19 classes into one virtual
class and rematch, which gives 84 test instances. A single class holds only
1--13 test instances, so per-class AP means nothing statistically. Tier-2 macro
AP50, the per-class mean, supplements it and is more sensitive to the hardest
classes.

\begin{table*}[t]
\centering
\caption{Downstream detection results on the Tier-2 rare classes at four
real-data fractions. Left: in-group pooled AP50 (main metric). Right: macro
AP50. Differences and ratios are computed from the unrounded values.}
\label{tab:detection}
\small
\setlength{\tabcolsep}{3pt}
\begin{tabular}{@{}lcccccc@{\hspace{1.6em}}ccccc@{}}
\toprule
& \multicolumn{6}{c}{\textbf{In-group pooled AP50 (main metric)}}
& \multicolumn{5}{c}{\textbf{Macro AP50}} \\
\cmidrule(lr){2-7} \cmidrule(lr){8-12}
\textbf{Fraction} & (i) Real & (ii) CP hard & (iii) CP Poisson & (iv) Diff.
& (iv)$-$best CP & (iv)/(i)
& (i) Real & (ii) CP hard & (iii) CP Poisson & (iv) Diff.
& (iv)$-$best CP \\
\midrule
10\%  & 0.0145 & \textbf{0.1108} & 0.1098 & 0.1073 & $-$0.003 (0.97$\times$) & 7.40$\times$
      & 0.007 & 0.042 & 0.029 & \textbf{0.045} & +0.003 (1.08$\times$) \\
20\%  & 0.0578 & 0.1528 & 0.1817 & \textbf{0.2775} & \textbf{+0.096 (1.53$\times$)} & 4.80$\times$
      & 0.047 & 0.140 & 0.109 & \textbf{0.173} & +0.034 (1.24$\times$) \\
50\%  & 0.1442 & 0.4390 & 0.3418 & \textbf{0.4738} & +0.035 (1.08$\times$) & 3.29$\times$
      & 0.116 & 0.321 & 0.276 & \textbf{0.393} & +0.071 (1.22$\times$) \\
100\% & 0.6722 & 0.8076 & 0.7923 & \textbf{0.8291} & +0.022 (1.03$\times$) & 1.23$\times$
      & 0.483 & 0.702 & 0.655 & \textbf{0.825} & \textbf{+0.122 (1.17$\times$)} \\
\bottomrule
\end{tabular}
\end{table*}

\textbf{Overall conclusion.} As the left panel of Table~\ref{tab:detection}
shows, diffusion is best on the main metric at the 20\%/50\%/100\% fractions.
It improves over the real-data-only baseline by $1.23\times$--$7.40\times$, and
its lead over the best Copy-Paste peaks at the 20\% fraction (+0.096,
$1.53\times$). Under the Tier-2 macro AP50 criterion it is best at all four
fractions (right panel). On the 14 head Tier-1 classes at the 100\% fraction the
four groups reach 0.966 / 0.954 / 0.970 / 0.963, a spread within
$\pm$0.008. The synthetic data does not harm the already saturated head
classes. The trend under mAP50-95 is the same. The last 8
epochs of val flatten out in every run, so under-training does not explain the
effect.

\textbf{Three advantages over the strongest Copy-Paste baseline.} First,
it works better on the hardest rare classes. Macro AP50 is best at
all four fractions (1.08--1.24$\times$), and the absolute lead at the 100\%
fraction reaches +0.122 against only +0.022 on pooled AP at the same fraction.
Macro weights every rare class equally, while pooling is dominated by
instance-rich classes. The gap between the two says the benefit concentrates on
the hardest classes. Second, it is best at
the medium scale across all four fractions, reaching 0.942 / 0.667 / 0.386 /
0.208 at the 100\%/50\%/20\%/10\% fractions. Third, direct template
pasting is not enough on complex symbol classes. Diffusion is markedly better
on the four classes \texttt{w34}/\texttt{w42}/\texttt{w24}/\texttt{w18}
(0.497--0.995 vs.\ 0.000--0.507), and its mean AP50 over the 12 target classes is
also the highest (0.770 vs.\ 0.606 for hard paste and 0.481 for Poisson). A
pasted patch is flat and adapts to no illumination, so its deviation from real
imaging grows with symbol complexity. Diffusion-generated symbols instead get
harmonised illumination, blur and noise.

\textbf{Two boundaries of applicability.} First, the pooled-AP advantage
does not widen monotonically as data shrinks. It peaks at the 20\%
fraction. At the 10\% fraction the three strategies converge under this
criterion: diffusion 0.107 vs.\ CP hard paste 0.111, a 0.003 gap below the
resolution that only 84 test instances afford. The training data is simply too
scarce there, and all three strategies sit close to the floor of this criterion.
Even at that fraction, diffusion is still best on macro AP and at
the small/medium scales. Second, small-object AP is not consistently
superior. Diffusion is best at the 100\%/10\% fractions, CP at
the 50\%/20\% fractions. Neither CP variant is consistently stronger than the
other: hard paste wins at 100\%/50\%, Poisson at 20\%, and they tie at
10\%. The comparisons above always take the stronger of the two per
fraction.

Diffusion thus brings a $1.23\times$--$7.40\times$ pooled AP50 improvement
over the real-data-only baseline. Copy-Paste reaches most of that improvement
too. The case for diffusion rests on the three advantages above, and the two
boundaries above bound it.

\section{Discussion and Conclusion}

\subsection{Failure Case Analysis}

\begin{figure}[t]
\centering
\includegraphics[width=\columnwidth]{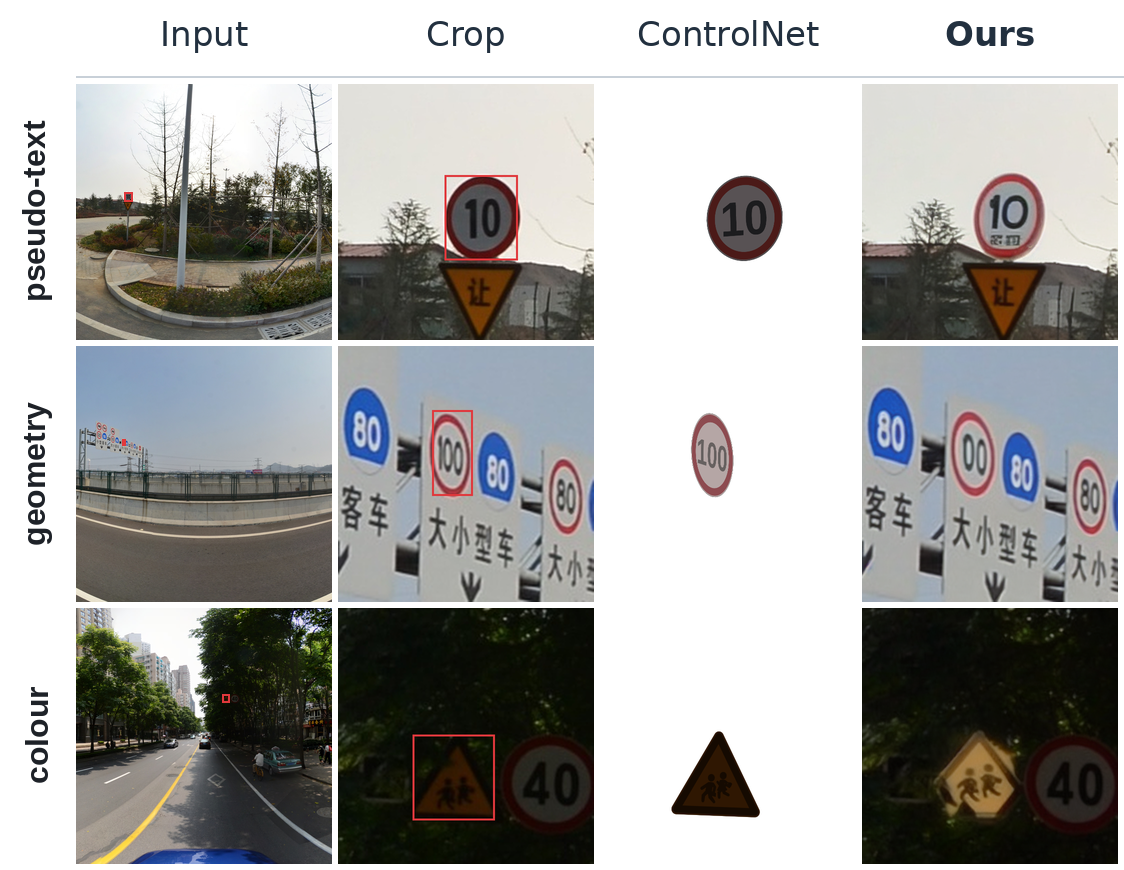}
\caption{Failure case grid. Columns: input, local crop (red box: the repaint
region), ControlNet guidance image, and our result. Rows: pseudo-text
generation, geometry--semantics conflict, and failure of colour clustering.}
\Description{Grid of three representative failure modes}
\label{fig:failure}
\end{figure}

The OCR confidence distribution separates cleanly, so we screen out the
low-confidence samples and summarise three typical failure modes
(Figure~\ref{fig:failure}).

\textbf{(1) Pseudo-text generation.} Real roads carry signs that do not fully
comply with the specification, for example a speed-limit sign with a line of
small annotation text beneath the digits. Once such samples enter the training
set, the model picks up the mistaken prior that a sign face may carry additional
text. It then adds pseudo-text of its own accord when generating a compliant
sign face.

\textbf{(2) Geometry--semantics tension from conflicting conditions.} Sometimes
the geometry of the ControlNet guidance image cannot host the layout the target
digits require, for instance a three-digit value on an extremely flattened
elliptical mask. The geometric constraint and the text generation then squeeze
each other. The digits come out flattened and deformed, sometimes with one digit
dropped altogether.

\textbf{(3) Failure of colour clustering.} A sign may be half covered by
shadow, or severely faded or soiled. The dominant colour from K-Means
clustering then departs from the colour scheme prescribed for that class, and
the injected reference colour guides the model towards a sign with a global
colour cast.

\subsection{Diversity and Potential Bias of the Synthetic Data}

\textbf{Sources and ceiling of diversity.} The diversity of our synthetic data
comes from three levels: the scene level (background, illumination, occlusion
and weather supplied by the original photograph), the pose level (imaged scale,
tilt angle and perspective supplied by the source instance), and the content
level (re-specification of the target category and digits). The first two levels
are inherited entirely from real data. The number and the distribution of the
source instances therefore set the ceiling of diversity.

\textbf{Three identified potential biases.}
(1) \textbf{A deliberate shift of the scale distribution.} Small objects make up
42\% of the generated set, above the 24--34\% of the natural distribution. We
introduced this on purpose to address the near-zero AP of small-object rare
classes under a real-data-only baseline. It does leave the scale distribution
of the synthetic data inconsistent with that of the test set.
(2) \textbf{An inherited bias in edit location.} Instances of a target class can
only appear at locations where a source sign once appeared. Some classes of sign
follow systematic placement regularities in the real world, merge signs at ramp
entrances for instance. If the source instances come from the locations of other
classes, the synthetic data introduces an erroneous ``location--category''
prior.
(3) \textbf{Risk of distribution shift induced by the quality gate.} Strict OCR
filtering biases systematically towards large-scale, high-contrast samples. We
relax the criterion to avoid this shift, intercepting only the case where the
value is read as another legal value. A small number of retained samples may
thus still carry an incorrect symbol, and warning-class samples are not
automatically verified at all.

\subsection{Conclusion}

This paper addresses the long-tailed data predicament in traffic sign detection
with a structured-prior-guided diffusion inpainting framework with physical
consistency. A traffic sign is a regulated artefact, defined precisely by a
national standard. We split its semantic, colour and geometric priors along
physical dimensions and inject each one in the conditioning form that matches
it. Semantics becomes structured JSON text. Appearance becomes a front-view
vector template painted with the measured dominant colour, injected via
IP-Adapter. Geometry becomes an affine-aligned template, injected via
ControlNet. A CIELAB chromaticity loss and a Sobel structure loss add two
physical consistency terms. Together with the leave-one-out protocol they
attribute each module's contribution to performance.

The experimental evidence covers three levels. On generation quality,
our method ranks first on all nine metrics in a cross-source zero-shot
evaluation over the full 5{,}182-sample TT100K set. Its OCR exact-match rate
reaches 91.1\%, 46.9 percentage points above the 12B FLUX.1 Fill, at $1/14$
of that model's inference time. On module contribution, leave-one-out ablations
quantify the decisive role of the appearance prior for digit legibility and of
colour injection for distributional realism. They also show that the Sobel loss
brings a structural gain at no distributional cost, and they report the
geometry--text trade-off of ControlNet openly. On the downstream task,
16 controlled training runs confirm that the synthetic data raises the pooled
AP50 of rare classes by $1.23\times$--$7.40\times$.

Three limits remain. Template coverage does not extend to non-circular and
non-triangular signs. Geometric fitting fails under large-angle occlusion. The
single-target editing framework runs serially. The downstream
conclusions apply to speed-limit classes and some warning signs, and the
marginal advantage over Copy-Paste tends to vanish at the 10\% data fraction.
Future work includes migrating to larger backbones such as FLUX, cross-validating
with a second detector, and extending laterally to inpainting tasks on other
regulated targets such as licence plates and road markings.

The framework assumes one thing: a specification defines the semantics,
appearance and geometry of the target precisely. Licence plates, road markings,
industrial nameplates and package labels satisfy that assumption too. Our route
therefore transfers to inpainting tasks on those regulated targets, and to
broader applications such as autonomous driving, street-view understanding and
automated high-definition map production.

\end{document}